\documentclass{article}

\usepackage[preprint]{neurips_data_2024}

\usepackage[utf8]{inputenc} 
\usepackage[T1]{fontenc}    
\usepackage{hyperref}       
\usepackage{url}            
\usepackage{booktabs}       
\usepackage{amsfonts}       
\usepackage{nicefrac}       
\usepackage{microtype}      
\usepackage{xcolor}         
\usepackage{graphicx}       
\usepackage{pdflscape}
\usepackage{float}
\usepackage{booktabs}
\usepackage{array}
\usepackage{multirow}
\usepackage{geometry}
\usepackage{makecell}
\usepackage{amssymb}
\usepackage{relsize}
\usepackage{subcaption}
\usepackage{amsmath}
\usepackage{ulem}
\usepackage{inconsolata}
\usepackage{tabularx}

\usepackage{listings}

\lstdefinestyle{prompt}{
    basicstyle=\ttfamily\small,
    backgroundcolor=\color{gray!10},
    frame=single,
    breaklines=true,
    postbreak=\mbox{\textcolor{red}{$\hookrightarrow$}\space},
    captionpos=b,
    numbers=none,
    columns=flexible,
    keepspaces=true
}

\usepackage{adjustbox}
\usepackage{array}

\usepackage{natbib}
\title{Meta-Models: An Architecture for Decoding LLM Behaviors Through Interpreted Embeddings and Natural Language} 
\author{%
  Anthony Costarelli\thanks{Equal contribution. Correspondence to: acostarelli@olin.edu and matdallen@gmail.com} \\
  Olin College of Engineering \\
  \And
  Mat Allen\footnotemark[1] \\
  Independent \\
  \And
  Severin Field \\
  Independent \\
}

\date{}

\begin{document}

\maketitle

\begin{abstract}
As Large Language Models (LLMs) become increasingly integrated into our daily lives, the potential harms from deceptive behavior underlie the need for faithfully interpreting their decision-making. While traditional probing methods have shown some effectiveness, they remain best for narrowly scoped tasks while more comprehensive explanations are still necessary. To this end,  we investigate meta-models—an architecture using a ``meta-model'' that takes activations from an ``input-model'' and answers natural language questions about the input-model's behaviors. We evaluate the meta-model's ability to generalize by training them on selected task types and assessing their out-of-distribution performance in deceptive scenarios. Our findings show that meta-models generalize well to out-of-distribution tasks and point towards opportunities for future research in this area. Our code is available at \url{https://github.com/acostarelli/meta-models-public}.

\end{abstract}

\begin{figure}[h!]
    \centering
    \includegraphics[width=0.8\textwidth]{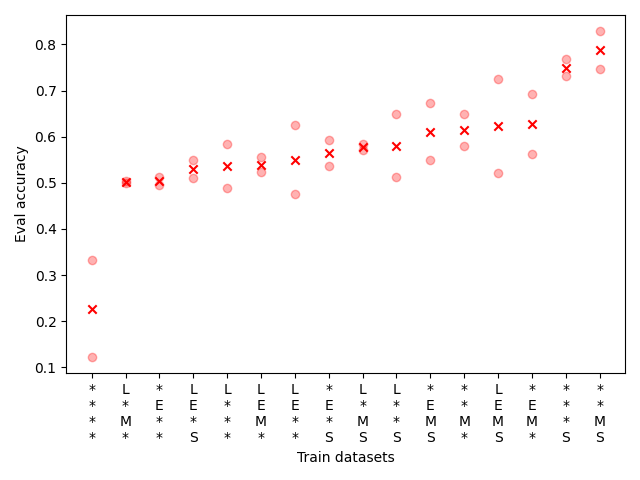}
    \caption{\textbf{Meta-model generalization accuracy} to lie detection when trained on datasets. Crosses are averages of scores across two runs. Datasets are \textbf{L}: Language, \textbf{E}: Emotions, \textbf{M}: Multilingual, \textbf{S}: Sentiment. Datasets and Results are discussed in depth in Section~\ref{sec:Dataset Utilization} and Section~\ref{sec:Empirical Findings} respectively.}
    \label{fig:main_figure}
\end{figure}

\section{Introduction}
AI systems continue to increase in capabilities and complexity, and are becoming more widely deployed in real-world use cases. As these systems continue to become responsible for more tasks in our own lives, the need becomes paramount to verify that a model will do what we intend it to do. Further, if an issue were to arise with that model, we would need to be able to see its inner workings to find out what went wrong in that system.

Currently, the human ability to understand decision-making within AI systems such as LLMs is at best limited, but showing promise \citep{templeton2024scaling}. Techniques such as probes allow us to look inside a model and see what it is “thinking” about when given a task \citep{gurnee2024languagemodelsrepresentspace}, and has shown efficacy in various domains \citep{hernandez2022ast, troshin2022probingpretrainedmodelssource}. Though powerful in some contexts, probes are not perfect \citep{Levinstein_2024}. In \citep{NEURIPS2022_725f5e80} the authors mention the unavoidable inaccuracies of probing classifiers' representations when they use ``non-concept features" and \citep{zhang-etal-2022-probing} points at the varying effectiveness when a model's input prompt changes. \citep{zhong2021factualprobingmasklearning} points at the impact that a base model's pretraining dataset has which allows for exploitation of probing accuracy. Most importantly, probes are limited in that they are trained for a specific task and offer little promise interpreting outside of their trained task set \citep{zhao2024explainability}.

Building off the lack of flexibility of probe-based interpretation methods, we extend research on the meta-model architecture \citep{langosco2024towards} which utilizes an existing LLM already trained to process to a wide variety of situations and fine-tune it to interpret the internal activations of another model. The setup takes the activations of one ``input-model'' along with a question about said activations and are fed to an ``interpreting'' model (referred here on as the ``meta-model''). The meta-model will then interpret and answer questions on the input-model's activations. This allows humans to interact with natural language questions about behaviors of the input-model with a ``natural language probe'' of sorts that has the potential to interpret a wide variety of behaviors.

An intended benefit of this approach is the ability to interpret a model’s internal processing and receive verification about decision-making that humans would not be able to make sense of based on output text alone. In other words, we would have a ``mind-reading'' AI model. For example, if a model is lying about a statement of which humans don't have access to the ground truth of that statement, we wouldn’t be able to verify the model's answer. But, if we were to instead rely on what's inside the ``brain'' of an LLM, we should receive more faithful results.

In this work, we specifically test the general interpretation abilities of the meta-model to extend to behaviors outside of its initial fine-tuning set. After fine-tuning the meta-model on initial task sets (e.g. detecting emotions, languages, etc.) we then evaluate its ability to interpret behaviors outside of that task set. In particular, we focus on interpreting a lying input-model.

We find that meta-models perform well in generalizing to interpreting lying in another model based solely on the input-model's internal activations. We also find that using models from different families still allows for interpreting behaviors with this architecture.

\section{Contributions}

In this work, we present several contributions to the field of AI Interpretability. Our contributions are as follows:

\begin{itemize}
    \item \textbf{Generalization of Meta-Models in Out-of-Distribution Tasks}: We further expand the meta-model architecture \citep{langosco2024towards} designed to interpret the internal activations of another LLM. We show an advantage over traditional probe techniques in the ability to generalize to new behaviors. 
    
    \item \textbf{Generalization Across Model Families}: We demonstrate that the meta-model architecture is capable of generalizing across different model families. Specifically, we show that a meta-model trained on \texttt{microsoft/phi-2} can effectively interpret behaviors in an unrelated input-model, \texttt{meta-llama/Llama-3.1-8B-Instruct}.
    
    \item \textbf{Evaluation of Deception Detection}: We test the meta-model's ability to generalize to detecting deceptive behavior in an input-model, even when trained on unrelated tasks. This shows the potential of the meta-model for identifying other safety behaviors in a robust manner.
\end{itemize}

\section{Related Works}
\subsection{Mechanistic Interpretability}
Mechanistic Interpretability seeks to make sense of the inner workings of the ‘black box’ of neural networks, with research spanning various levels of abstraction. Early work \citep{rumelhart1986learning} recognized that hidden units in neural networks could represent important features. The field has examined neural networks through different lenses: scrutinizing individual neurons \citep{alain2016understanding}, mapping connections between neurons in circuits \citep{olah2020zoom}, analyzing weights to determine model characteristics \citep{Krizhevsky2009LearningML, Ha2016HyperNetworks}, and uncovering high-level model behaviors and representations \citep{gurnee2024languagemodelsrepresentspace, li2022emergent, nanda2023emergentlinearrepresentationsworld}. While much work has been done in this area, Mechanistic Interpretability is a young field with many unsolved problems \citep{bereska2024mechanisticinterpretabilityaisafety, elhage2022toy}. Given the limitations of the field, we explore a more flexible approach that leverages the power of LLMs to interpret neural activations. With the meta-model architecture, we can interpret any prompt from one model and ask any number of questions about it to receive natural language descriptions of said model.

\subsection{Automated Interpretability}
Automated Interpretability arises in response to the complexities inherent in manual interpretability methods even when applied to smaller, simplified models \citep{elhage2021mathematical, elhage2022toy}. While some automated techniques like Sparse Autoencoders \citep{bricken2023towards} or automated interpretability agents \citep{shaham2024multimodalautomatedinterpretabilityagent, schwettmann2023findfunctiondescriptionbenchmark} have eased efforts and found interesting results \citep{templeton2024scaling, cunningham2023sparseautoencodershighlyinterpretable, gurnee2023findingneuronshaystackcase}, these methods differ from our work in that they focus on interpreting specific neurons where we focus on higher abstractions of behavior.

\subsection{Meta-Models}
Meta-models, a subfield of Automated Interpretability, leverages one model’s decoding capabilities to interpret high level details of another \citep{langosco2024towards}. An early exploration of this concept was conducted in \citep{bills2023language}, where a GPT-2 neuron's activations were formatted as a prompt for GPT-4 to generate an explanation. One significant meta-model approach uses classifiers to determine ``internal'' states of models, as demonstrated in \citep{kolouri2020universallitmuspatternsrevealing, xu2020detectingaitrojansusing, langosco2024towards}. Another example is the use of an autoencoder trained on the weights of many models to infer behaviors \citep{schürholt2022hyperrepresentationsselfsupervisedrepresentationlearning}. In this work, we aim to interpret one model’s activations using another model’s decoding abilities, without needing to train a model from scratch.

\subsection{Activation Patching}
Activation Patching, a recent technique for analyzing and manipulating neural networks, involves reading or selectively modifying (or ``patching'') specific activations within a model to either interpret or modify its behavior. Decoding the intermediate representations allows us to understand the progression of decision-making in a model \citep{nostalgebraist2020interpreting}. Building on this paradigm, by interpreting representations in a network we can also edit specific representations to alter factual knowledge \citep{meng2023locatingeditingfactualassociations}, aid in safety tasks \citep{turner2024activationadditionsteeringlanguage, zou2023representationengineeringtopdownapproach}, or even fine-tune models \citep{chen2024selfieselfinterpretationlargelanguage} based on the model's internal representations. Our work diverges from these approaches by focusing solely on interpreting existing behaviors without modifying the observed model. The closest related studies to our tasks are \citep{ghandeharioun2024patchscopesunifyingframeworkinspecting, chen2024selfieselfinterpretationlargelanguage}, who similarly take one model’s activations and patch them into another for interpretation. Though we build off of \citep{chen2024selfieselfinterpretationlargelanguage} in this study (see \autoref{sec:meta-model-architecture} for implementation details), we instead explore the ability of a meta-model to generalize its ability to interpret another model's behaviors from its activations. To the best of our knowledge, no comprehensive study has been conducted on the generalizability of a meta-model architecture in interpreting and reporting behaviors in out-of-distribution tasks, which is the primary focus of our research.

\section{Methodology}

In this section, we detail the architecture and implementation of our experiments. Our setup is based on \citet{chen2024selfieselfinterpretationlargelanguage} and involves conditioning an input-model to exhibit specific behaviors, capturing its hidden activations, and using the meta-model to interpret the input-model's activations to determine those specific behaviors.

\subsection{Dataset Utilization}
\label{sec:Dataset Utilization}
Each dataset is employed to elicit specific behaviors from the input-model by crafting tailored prompts (e.g. \lstlistingname~\ref{lst:Example Input-Model Prompt}). Each dataset is balanced such that $\mathrm{P}(\mathrm{Yes} | \mathrm{question})=0.5$ for every question.

\begin{itemize}
    \item \textbf{Sentiment Analysis (SST-2) \cite{wang2019glue, socher2013recursive}}: 67.3k movie reviews labeled as positive or negative, used for binary sentiment classification.
    
    \item \textbf{Emotion Detection (GoEmotions) \cite{demszky2020goemotions}}: Contains 58k Reddit comments annotated with 28 distinct emotions, used to classify emotional behaviors.
    
    \item \textbf{Language Identification \cite{keung-etal-2020-multilingual}}: This dataset includes text samples in six languages—English, Japanese, German, French, Spanish, and Chinese.
    
    \item \textbf{Multilingual Sentiment Analysis \cite{keung-etal-2020-multilingual}}: We make use of the same Language Identification dataset as above and incorporate its sentiment labels to test on sentiment classification across multiple languages.
    
    \item \textbf{Lying Detection \cite{pacchiardi2024how}}: A set of 380 ``questions to which the model has not encountered the answer during pre-training (e.g., a statement about a fictional character) but for which we provide the correct answer in the prompt.''  We similarly provide the correct answer in the input-model's prompt and also use the author's definition of what it means for a model to lie; that it is ``outputting false statements despite `knowing' the truth in a demonstrable sense'' \cite{pacchiardi2024how}.
\end{itemize}

\subsection{Meta-Model Architecture}
\label{sec:meta-model-architecture}

The Meta-Model Architecture relies on two models: the input-model and the meta-model. For the input-model we use \texttt{meta-llama/Llama-3.1-8B-Instruct} \citep{dubey2024llama3herdmodels} and for the meta-model we use \texttt{microsoft/phi-2} \citep{javaheripi2023phi}. Both are accessed via the Transformers Hugging Face API \citep{wolf2020huggingfacestransformersstateoftheartnatural}.

The meta-model is trained to classify the input-model behaviors by being given a ``Yes'' or ``No'' question and the captured internal activations of the input-model. To generate the meta-model's prediction it is prompted with a question and a set of placeholder tokens (e.g. \autoref{lst:Example Meta-Model Prompt}). The placeholder tokens are swapped with the internal activations immediately after the embedding layer. We perform no finetuning of the input-model.

Our methodology consists of four main steps:

\clearpage
\begin{enumerate}
    \item \textbf{Conditioning the Input-Model:} We employ few-shot prompting to elicit specific behaviors from the input-model. For instance, to condition the model to exhibit negative sentiment, we structure the prompt as follows:
    \begin{lstlisting}[style=prompt, label=lst:Example Input-Model Prompt, caption=Example Input-Model Prompt]
        user: What did you think of the movie?
        input-model: It was terrible!
        user: What else?
        input-model: {another negative example}
        user: Say more.
        input-model: {another negative example}
        user: Keep going.
        input-model:
    \end{lstlisting}

    \item \textbf{Capturing Internal Activations:} During a single forward pass of the conditioned input-model, we capture the internal activations from selected layers $L_1, L_2, \dots, L_n$. These activations are stored as vectors $\mathbf{A}_1, \mathbf{A}_2, \dots, \mathbf{A}_n$.

    \item \textbf{Prompting the Meta-Model:} The meta-model is prompted with a ``Yes'' or ``No'' question regarding the input-model's behavior, alongside placeholder tokens that will be replaced by the captured activations after the embedding layer (as seen in ~\autoref{lst:Example Meta-Model Prompt}):
    \begin{lstlisting}[style=prompt, label=lst:Example Meta-Model Prompt, caption=Example Meta-Model Prompt]
        user: Is this model acting negative? <placeholder1><placeholder2><placeholder3> 
        meta-model: {Yes/No}
    \end{lstlisting}

    \item \textbf{Intercepting and Replacing Placeholders:} During the meta-model's forward pass, we intercept the placeholder tokens and replace them with the captured activations $\mathbf{A}_1, \mathbf{A}_2, \dots, \mathbf{A}_n$. The meta-model then continues operating as normal. 
\end{enumerate}

\section{Empirical Findings}
\label{sec:Empirical Findings}

\subsection{Generalization to lie-detection}

Our main results are located in \autoref{fig:main_figure}. The results strongly suggest that the meta-model architecture effectively promotes generalization to our designated out-of-distribution task of detecting lying. The steady improvement in performance as different dataset combinations are incorporated indicates that the model can learn to interpret information about a given input-model to enhance its predictive capabilities. This finding opens up numerous research opportunities which are listed in \autoref{sec:future_work}.

\subsubsection{Best and worst performance}
The poorest performance is observed with the untrained meta-model, achieving an average evaluation accuracy of about 0.2, showing a drastic improvement of any training at all. In contrast, the highest accuracies were from models trained on the Sentiment and/or Multilingual dataset alone. Performing more experiments may show different performance, but ``more'' didn't necessarily mean ``better''; possibly suggesting destructive interference from different datasets. For example, the model trained on Sentiment alone scored second highest with 0.75 but the model trained on Sentiment, Emotions, and Language identification was just above chance at 0.5.

\subsubsection{Impact on performance from datasets}
The Multilingual dataset consistently showed a positive impact on performance when included in training combinations. Conversely, while the Emotions and Language datasets sometimes contributed to improved performance, in other cases it appeared to have a neutral or slightly negative effect. 

There's a general trend of improved performance as more datasets are combined, but this is not strictly linear. Some combinations perform better than others, suggesting complex interactions between datasets. It's worth noting that the model trained on all datasets was the fourth highest rating, where the model trained only on Sentiment was second highest with an approximate 0.10 margin.

\subsubsection{Performance on other models}

\begin{figure}[h!]
    \centering
    \includegraphics[width=0.8\textwidth]{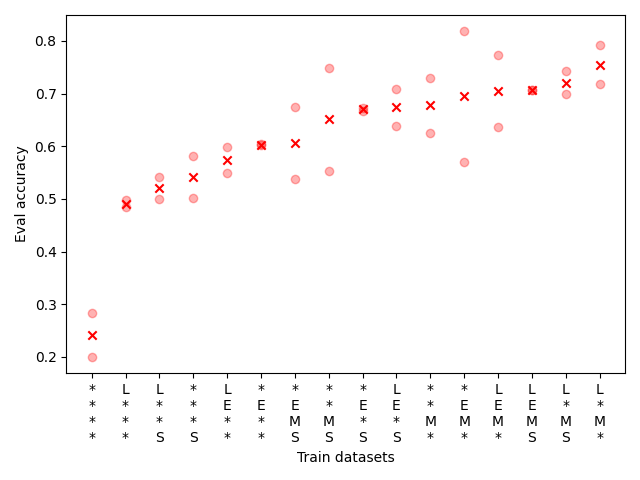}
    \caption{\textbf{InternLM as meta-model} Crosses are still averages of scores across two runs. Entire experiment setup used on \texttt{meta-llama/Llama-3.1-8B-Instruct} are recreated exactly here.}
    \label{fig:internLM_results}
\end{figure}

We extended our experiments to the \texttt{internlm/internlm2\_5-7b-chat} \citep{cai2024internlm2} model to further assess the generalization to other models. The results listed in Figure \ref{fig:internLM_results}, show several similarities with the \texttt{meta-llama/Llama-3.1-8B-Instruct} experiments while also showing unique results.

As in the main experiment, the Multilingual dataset continued to show a strong positive effect on model performance. Configurations that included the Multilingual dataset consistently outperformed those that did not. Most notably, the two highest and lowest performing (outside of `no training') configurations are a difference of Multilingual being added or not. Language and Language + Sentiment scored lowest but with Multilingual added they were the highest scores.

Similar to the trends observed in the \texttt{meta-llama/Llama-3.1-8B-Instruct} experiments, we saw a general increase in accuracy as more datasets were incorporated into the training. Again, however, the improvement was not strictly linear. Strikingly different from the main results is how \texttt{internlm/internlm2\_5-7b-chat} saw an increase from the Language dataset, where the main experiments showed worse performance when trained on this dataset. This further shows that datasets have a more complex relationship than ``more is better''.

In general, the findings from the \texttt{internlm/internlm2\_5-7b-chat} experiments strongly align with the results observed in the \texttt{meta-llama/Llama-3.1-8B-Instruct} model, demonstrating further robustness of the meta-model architecture.

\section{Future Work}
\label{sec:future_work}

Our findings open up many future research directions. We outline some areas that warrant further investigation to expand on the meta-model approach.

\begin{itemize}
    \item Investigating the optimal combination and sequencing of datasets for training
    \item Experimenting with different parameter sizes, layer sampling, number of layers, and spread of sampling points in the input-model. We take samples of one token every four layers, would wider spreads reveal different information?
    \item Testing on further out-of-distribution safety relevant tasks (in addition to lying, detecting malware generation, sycophancy, backdoor detection, etc.)
    \item Explore the causal effect of datasets on generalization abilities and training on new datasets, noting if there's a constructive or destructive influence from a dataset on a model's performance
    \item Generating and training on new types of datasets
    \item The effect of datasets on different types of models. For example, Language performs poorly with \texttt{meta-llama/Llama-3.1-8B-Instruct} but performs well (\autoref{fig:internLM_results}) with \texttt{internlm/internlm2\_5-7b-chat} 
    \item Influence the effect of training meta-models to output ``yes'' or ``no'' before any other training
    \item Training an instruct-meta-model to give a greater depth of description of the input-model
    \item Extend to multi-modal models and determine generality
    \item Interpret a model during its pretraining phase 
    \item Identify adversarial attacks against meta-models
    \item Training input-models to be more interpretable, possibly through LoRA layers
\end{itemize}

\section{Conclusion}

Our study demonstrates the effectiveness of meta-models in generalizing to out-of-distribution model interpretation, focusing particularly on detecting deception without needing a given model's output text. We also demonstrate the efficacy across different model families using \texttt{microsoft/phi-2} as the meta-model and \texttt{meta-llama/Llama-3.1-8B-Instruct} as the input-model. We find that meta-models offer superior generalization to out-of-distribution tasks compared to traditional probing methods. Though it remains unclear the exact effect that training on different combinations of datasets provides, training the models on sentiment, emotion, language identification, and multi-lingual sentiment tasks does increase performance in detecting deception.

This method of interpreting model behavior without relying on output text also opens up new possibilities for real-time monitoring and auditing of AI systems when we're not certain we can trust a model's output text. This research takes a step towards demystifying AI decision-making processes and extends new directions for researchers to explore more advanced meta-model architectures and training strategies.

\appendix
\section{Acknowledgments}
This work was graciously supported by a grant from EA Funds.

We would like to thank Joshua Clymer for developing the project proposal, scope, and weekly advising meetings.
\bibliography{citations}

\end{document}